\documentclass[10pt,twocolumn,letterpaper]{article}

\usepackage{wacv}              

\usepackage[accsupp]{axessibility} 
\usepackage{graphicx}
\usepackage{amsmath}
\usepackage{amssymb}
\usepackage{booktabs}
\usepackage{enumitem}
\usepackage{tabularx}

\usepackage[pagebackref,breaklinks,colorlinks]{hyperref}

\usepackage[capitalize]{cleveref}
\crefname{section}{Sec.}{Secs.}
\crefname{section}{Section}{Sections}
\crefname{table}{Table}{Tables}
\crefname{table}{Tab.}{Tabs.}

\def\wacvPaperID{\vspace{=1cm}2522} 
\def\confName{WACV}
\def\confYear{2024}

\begin{document}
\newcommand{\oursnospace}{U3DS$^3$}
\newcommand{\ours}{U3DS$^3$ }
\newcommand{\Mod}[1]{\textcolor{blue}{#1}}
\newcommand{\Ori}[1]{\textcolor{red}{#1}}

\title{\vspace{-1cm}\oursnospace: Unsupervised 3D Semantic Scene Segmentation \vspace{-0.6cm}}


\author{Jiaxu Liu$^{1}$\qquad Zhengdi Yu$^{1}$\qquad Toby P. Breckon$^{1,2}$\qquad Hubert P. H. Shum$^{1}$\\
        Department of \{Computer Science$^{1}$ \textbar 
 \space Engineering$^{2}$\}, Durham University, UK \\
        \{\tt\small jiaxu.liu, zhengdi.yu, toby.breckon, hubert.shum\}@durham.ac.uk
}

\maketitle

\begin{abstract}
Contemporary point cloud segmentation approaches largely rely on richly annotated 3D training data. However, it is both time-consuming and challenging to obtain consistently accurate annotations for such 3D scene data. Moreover, there is still a lack of investigation into fully unsupervised scene segmentation for point clouds, especially for holistic 3D scenes. This paper presents \oursnospace, as a step towards completely unsupervised point cloud segmentation for any holistic 3D scenes. To achieve this, \ours leverages a generalized unsupervised segmentation method for both object and background across both indoor and outdoor static 3D point clouds with no requirement for model pre-training, by leveraging only the inherent information of the point cloud to achieve full 3D scene segmentation. The initial step of our proposed approach involves generating superpoints based on the geometric characteristics of each scene. Subsequently, it undergoes a learning process through a spatial clustering-based methodology, followed by iterative training using pseudo-labels generated in accordance with the cluster centroids. Moreover, by leveraging the invariance and equivariance of the volumetric representations, we apply the geometric transformation on voxelized features to provide two sets of descriptors for robust representation learning. Finally, our evaluation provides state-of-the-art results on the ScanNet and SemanticKITTI, and competitive results on the S3DIS, benchmark datasets.
\end{abstract}


\section{Introduction}
\noindent
As a crucial task in 3D computer vision, there has been increasing attention paid to point cloud segmentation in recent years due to its broad applicability to many real-world applications such as autonomous driving, virtual reality, robotics, and human-computer interaction. However, owing to the unordered and unstructured nature of point clouds, it is a non-trivial exercise to undertake segmentation upon them. In recent years, supervised point cloud segmentation approaches have made significant progress \cite{qi2017pointnet,pointnet++, kpconv, rand, dgcnn,Li_2023_CVPR,df} against several benchmark datasets \cite{s3, scannet,shapenet2015, behley2019semantickitti}. However, these approaches rely heavily on copious fully-annotated training data, in the form of labeled 3D point clouds. It is both time-consuming and labour-intensive to obtain such annotations accurately and consistently - especially for dense and complex 3D scenes.  An alternative body of work leverages semi-supervised \cite{semi} and weakly-supervised \cite{weakly,weakly-EFEM,weakly-box2mask,weakly-dataefficient} approaches to mitigate the labelled data requirements, but still require labour-intensive annotation at some level and lack of being readily scalable and adaptable to new datasets. Our work aims to characterize 3D features without any explicit guidance allowing it to learn from the intrinsic structure of the data, and offer independence from erroneous, bias or inconsistent annotations, which significantly differ from prior weakly-supervised methods. To date, there are only a handful of prior works trying to address fully unsupervised segmentation for point clouds \cite{Unsup,co-seg, ogc, unsupervised3DGANs}. However, these approaches essentially focus on object-level segmentation or co-segmentation and cannot recover the full 3D scene labels without extra scene priors \cite{co-seg, ogc, unsupervised3DGANs} and only a recent work \cite{Unsup} has attempted to address fully unsupervised semantic segmentation for 3D scenes. Our proposed new \ours approach performs full holistic segmentation for the entire 3D scene in a scene-agnostic manner, spanning both indoor and outdoor scenarios across differing metric scales and achieving superior results on ScanNet\cite{scannet} and SemanticKITTI\cite{behley2019semantickitti} when compared to \cite{Unsup}.

\begin{figure}[t]
\begin{center}
\includegraphics[width=1.0\linewidth,height=0.5\linewidth]{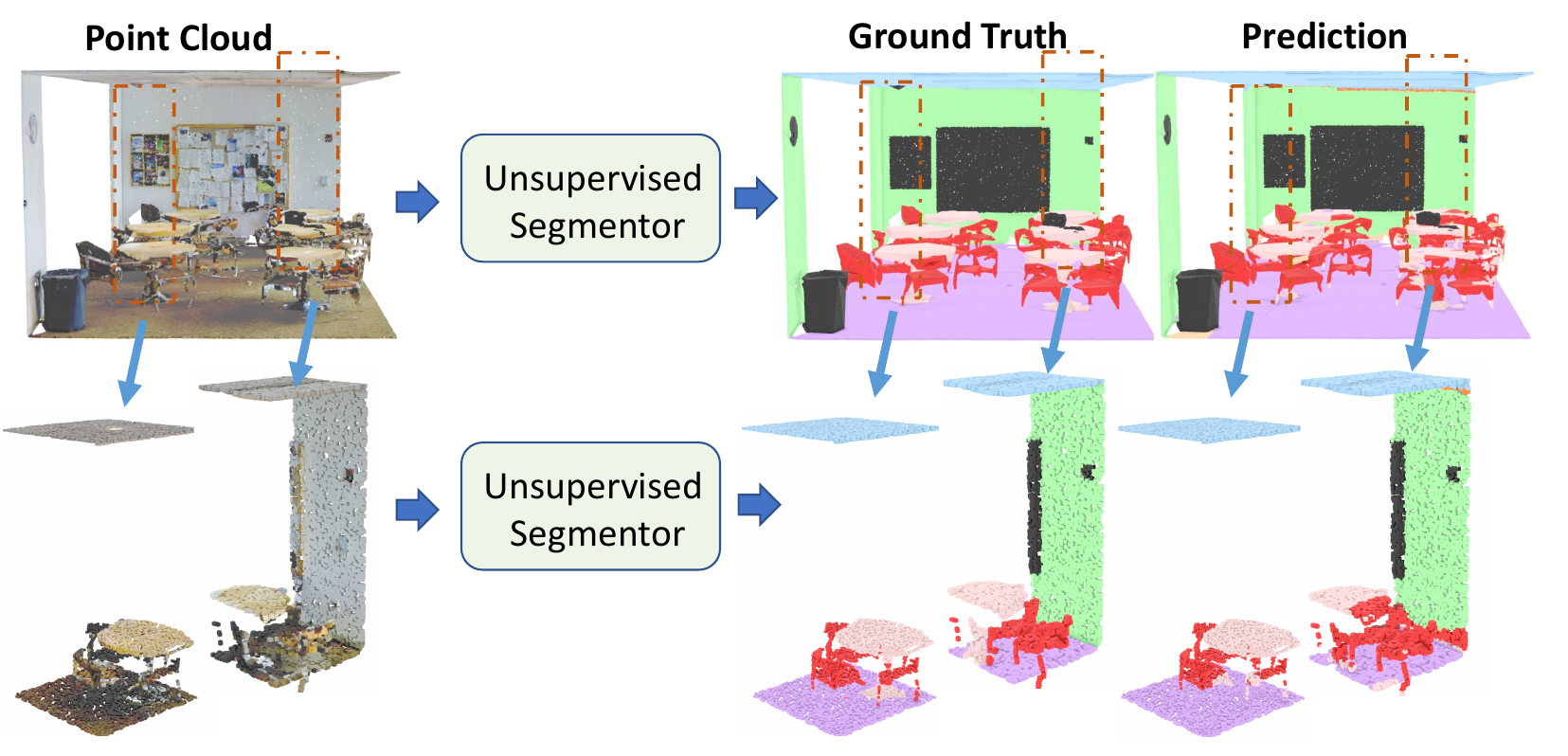}
\end{center}
   \vspace{-6mm}
   \caption{\ours unsupervised point cloud semantic segmentation (illustrated on S3DIS dataset \cite{s3}). Left to right: real scene, ground truth and \ours segmentation results for the full scene (upper), for a single point cloud block input (lower).}
   \vspace{-2mm}
\label{fig:open}
\end{figure}

Despite the growth of unsupervised learning on 2D image segmentation \cite{picie, iic, ac,2DULcontrasmask,2dULGans}, there is a lack of in-depth investigation into any 3D point cloud equivalent. Although some achievements in unsupervised segmentation learning have addressed 3D point cloud data via domain adaptation \cite{domain1,domain2}, our work does not rely upon transfer learning. OGC \cite{ogc} leverages the dynamic motion pattern of a (LiDAR derived) point cloud sequence to acquire dynamic tracks and achieve competitive results for object-level segmentation. Similarly, Yang et al. \cite{co-seg} successfully apply unsupervised learning for object co-segmentation in point clouds. \cite{Unsup} made the first attempt towards unsupervised 3D semantic segmentation via region growing to generate high-quality over-segmentation, but their method does not fully leverage the intrinsic geometric information of the point clouds and tends to predict over-smooth segmentations with more background (e.g. floor, wall) and overlook detailed object categories of the scene.

Traditional clustering methods, like \textit{k}-means \cite{KMeans} and DBSCAN \cite{DBSCAN}, can be beneficial in establishing unsupervised semantic segmentation baselines. However, these methods still exhibit notable drawbacks. \textit{k}-means \cite{KMeans}, for instance, struggles to converge effectively with non-convex datasets, exhibits weaknesses in handling uneven data distributions, and struggles to form coherent clusters in the presence of outliers and data noise. Interestingly, some existing unsupervised approaches \cite{picie,Unsup} also incorporate \textit{k}-means as a component of their algorithms. On the other hand, DBSCAN \cite{DBSCAN} encounters challenges when dealing with categorical features, often fails to identify clusters with varying densities, requires a drop in density to identify boundaries, and experiences decreased performance in high-dimensional scenarios.

The goal of our approach is to enable a generalized method that is able to perform semantic segmentation for large-scale indoor and outdoor 3D scenes without utilizing any human labels or dynamic information between LiDAR frames. This paper takes a new step towards scene-level unsupervised semantic segmentation with a novel strategy. Specifically, we first apply voxel cloud connectivity segmentation (VCCS)\cite{VCCS} to generate the initial superpoint and merge them according to the distance and normals of the superpoints. Following this, we propose the baseline method by applying mini-batch \textit{k}-means \cite{mini-batch} on the features of a 3D point cloud to generate and update the clustering centroids, and subsequently calculate the distance between features and clustering centroids to assign labels for each point as pseudo-labels under the guidance of the superpoint. After that, we train the network with the pseudo-labels to provide new network parameters for the next iteration of clustering. Subsequently, we apply a non-parametric classifier that operates solely on the feature space distance. Finally, by leveraging the invariance and equivariance of the volumetric representations, we are able to apply differing volumetric transformations on the point cloud input and a subsequent voxelized reverse geometric transformation on these feature representations.  

In this manner, our network is capable of producing several variant feature representations from the same data source. This transformation operation is derived from a very intuitive sense that the same inputs should result in similar predictions even under geometric transformation due to the principle of invariance. Fundamentally, we learn a feature representation that maximizes the effective semantic class separation. We provide two pathways to enforce color invariance and geometric equivariance that each provide our underlying inductive bias for semantic consistency and geometric structure by way of consistent clustering assignment across the two pathways. This is performed via iterative optimization of the clustering loss, which enforces a discriminative feature space capable of high-level visual similarity disambiguation. Finally, we train our voxel-based method in an end-to-end manner. Furthermore, our evaluation illustrates promising results across both indoor and outdoor datasets, S3DIS \cite{s3}, ScanNet\cite{scannet} and SemanticKITTI \cite{behley2019semantickitti}, demonstrating the effectiveness and practicality of our method and providing an initial reference performance for completely unsupervised 3D semantic scene segmentation.  Overall, we propose a simple yet effective framework that makes the new approach towards the task of unsupervised point cloud segmentation for holistic 3D scenes, named \textbf{\oursnospace}. \cref{fig:open} illustrates an initial qualitative result of our approach. Our key contributions are summarized as: 

\begin{itemize}[topsep=0.5pt, partopsep=0pt, parsep=0.3pt, itemsep=0.3pt]
    \item We propose a novel unsupervised semantic segmentation method to leverage the invariance and equivariance through geometric transformation for both 3D indoor and outdoor holistic scenes.
    \item We analyze and compare existing clustering approaches and the concurrent state-of-the-art, demonstrating the advantages and superiority of our method for efficient unsupervised learning on large-scale point clouds of holistic 3D scenes with faster convergence.
    \item We conduct extensive experiments and ablation studies to demonstrate significant improvement over standard baselines, across the S3DIS \cite{s3} ScanNet\cite{scannet} and SemanticKITTI \cite{behley2019semantickitti} benchmark datasets, and illustrate both the practicability of the proposed framework and justify the intuition behind our design.
 
\end{itemize}


\section{Related Work}
\noindent
In this section, we briefly summarize the prior literature on 3D Semantic Segmentation (\Cref{sec:3d-sec-lit-review}) and Unsupervised Segmentation Learning (\Cref{sec:unsup-sec-lit-review}).

\begin{figure*}[hbt!]
\begin{center}
\includegraphics[width=0.9\linewidth]{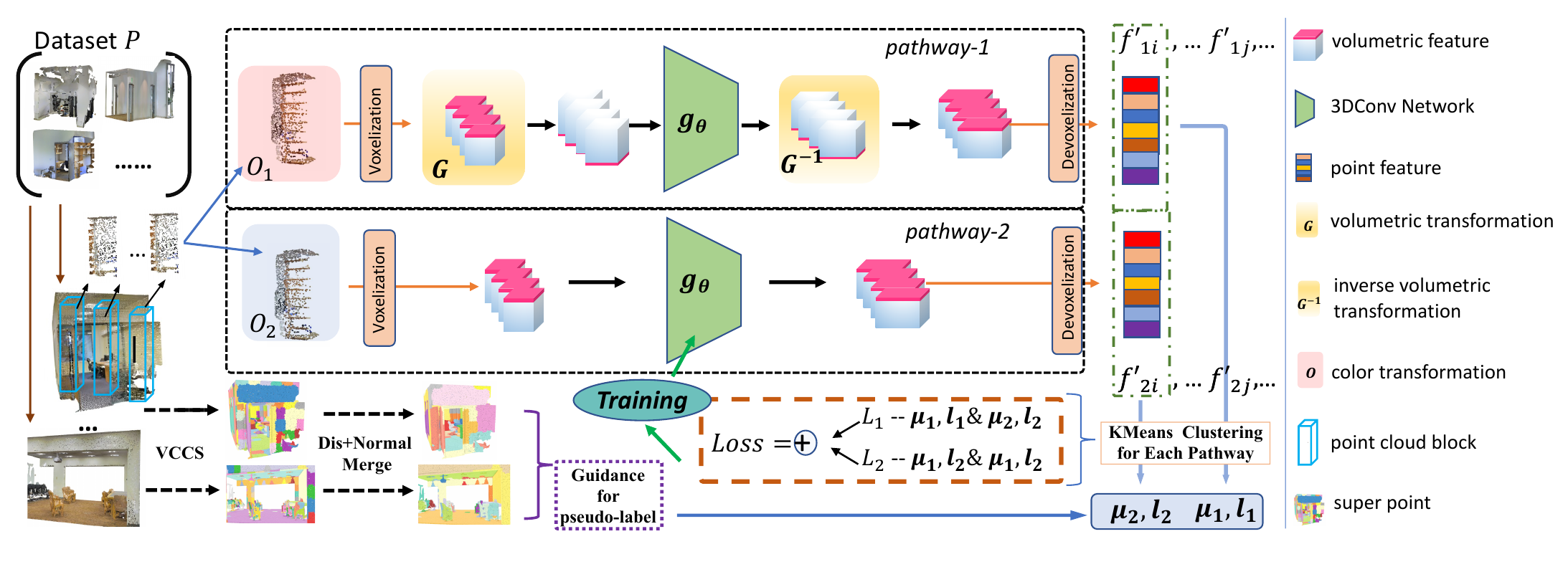} 
\end{center}
   \vspace{-7mm}
   \caption{The illustration of the proposed unsupervised semantic segmentation method. Each input point cloud is assigned to two pathways and gives two groups of clustering centroids and labels for training. The point cloud is initially calculated to form a superpoint. This superpoint is then merged to produce a refined superpoint, which guides the generation of pseudo-labels. The pink part of the volumetric feature here denotes reverse the input tensor along the $z$ axis.}
   \vspace{-2mm}
\label{fig:overview}
\end{figure*}


\subsection{3D Semantic Segmentation} 
\label{sec:3d-sec-lit-review}
\noindent  
To learn per-point semantics for 3D point clouds, many deep learning based approaches tackle 3D point cloud semantic segmentation tasks. PointNet \cite{qi2017pointnet} is a pioneering work and the first one to leverage the point-based encoding strategy, which is able to directly learn point features from the raw points and extract local information embedded in the neighbouring points. Following this work, more point-based methods \cite{pointnet++, kpconv, rand} have been proposed. KPConv \cite{kpconv} designs a kernel function for operating convolution in 3D space to tackle local geometric structures. RandLA-Net \cite{rand} proposes a more efficient framework by replacing a complicated point selection strategy with random sampling. On the other hand, voxel-based approaches \cite{FCPN,voxelnet} typically employ a 3D convolutional neural network by converting the point cloud from uneven distribution to regular voxel grids. Some works \cite{sparse4d,sparseconv} explore the more efficient voxel-based method by conducting sparse convolution. PVCNN \cite{pvcnn} proposes a point-voxel corporate method and utilizes trilinear de-voxelization for voxelized features for fine-grained feature extraction. Our method also leverages a 3D convolutional neural network and follows the trilinear de-voxelization approach from \cite{pvcnn} to avoid extracting identical features for the points that lie in the same voxel grid. Moreover, \cite{graph1,graph2} bring in graph convolutions to learn point features. However, all of these fully supervised methods require richly annotated training data, which are time-consuming and labour-expensive to obtain. To address the issue, Jiang et al. \cite{semi} propose a semi-supervised contrastive learning method for alleviating the tedious labelling cost. Zhang et al. \cite{weakly} utilizes perturbed self-distillation to employ a weakly supervised method for point cloud semantic segmentation and reducing human annotations. In terms of unsupervised manner, KMeans \cite{KMeans} and DBSCAN \cite{DBSCAN} are classic methods and have no requirement for labelled data. However, these methods can only deal with simple object-level segmentation and lack of robustness under non-convex and uneven data distribution. 

\subsection{Unsupervised Segmentation Learning} 
\label{sec:unsup-sec-lit-review}
\noindent
The exploration of unsupervised 2D semantic segmentation shows more maturity compared to 3D. DeepCluster \cite{deepcluster} clusters the feature vector of the entire dataset using \textit{k}-means \cite{KMeans} to assign pseudo-labels and subsequently trains its encoder. Our method shares the common idea that iterative optimization of clustering can improve feature representation learning. Abdal et al. \cite{2dULGans} propose an unsupervised segmentation framework that enables foreground and background separation for raw input images and segments class-specific Style-GAN images. Liu et al. \cite{2dULpartseg} study the segmentation for object parts by means of semantic consistency of object parts, where the segmentation regions of the same part should be semantically consistent across object instances and robust to appearance and shape changes. Gansbeke et al. \cite{2DULcontrasmask} put forward a two-step framework that adopts a predetermined mid-level prior in a contrastive optimization objective to learn pixel embedding for semantic segmentation. PICIE \cite{picie} notably proposes a pixel-level semantic segmentation method that can incorporate geometric consistency as an inductive bias to learn invariance and equivariance for photometric and geometric variations. 

As for the 3D domain, some work already shows notable performance on unsupervised segmentation, but not especially for scene-level semantic segmentation. Yang et al. \cite{co-seg} employs object sampler and background sampler to tackle unsupervised point cloud co-segmentation for object-level segmentation by co-contrastive learning and mutual attention sampling. However, the co-segmentation method only works for objects and it is limited by the groups of common 3D objects. Some other works \cite{domain1,domain2} focus on unsupervised domain adaptation for point cloud semantic segmentation. OGC \cite{ogc} can simultaneously identify multiple 3D objects in a single forward pass, without any human annotations, which leverages the dynamic motion patterns over (LiDAR captured) sequential point clouds as supervision signals to automatically discover rigid objects. However, OGC \cite{ogc} needs the dynamic information of continuous point cloud frames as an input prior. Poux \textit{et al}. \cite{3DUL} leverages the region growing method for indoor unsupervised object-level segmentation, but the segmentation is only for generating larger object segment parts. GrowSP\cite{Unsup} is the only unsupervised 3D semantic segmentation that employs a progressively region-growing scheme to generate high-quality over-segmentation, however, their method does not fully leverage the intrinsic geometric information of the point clouds and tends to predict over-smooth segmentation and lose accuracy in intricate scenes. In contrast, our work aims to investigate unsupervised 3D semantic segmentation leveraging the intrinsic geometric information of the point clouds for holistic 3D scenes without any dynamic information or transfer learning prior.


\section{{\ours} Methodology}
\noindent
 This work formulates the task of unsupervised point cloud semantic segmentation as point-level segmentation, where every point within the point cloud needs to be assigned a label of a fixed number of semantic class labels.

To state formally, given a point cloud set $\boldsymbol{P}$ without labels, let $\boldsymbol{c} = \left\{c_{i}\right\}$ and $\boldsymbol{f} = \left\{f_{i}\right\}$ denote the point coordinates and the corresponding features from
$\boldsymbol{P}\in \mathbb{R}^{N\times 3}$, $\boldsymbol{F}\in \mathbb{R}^{N\times d}$, where $N$ is the number of points of the input point cloud, and ${d}$ denotes the feature size, which contains coordinates, colours, and normalized positional information. Hence, the goal of this work is to learn a semantic segmentation function  $\boldsymbol{g}_{\theta }$, which is able to predict per-point labels in an unsupervised way for $\boldsymbol{P}$ using only $\boldsymbol{c}$ and $\boldsymbol{f}$.

As shown in \cref{fig:overview}, for each input point region, we first apply two different colour transformations and afterwards convert them to the volumetric domain. For \textit{pathway-1} in the top row, we implement a geometric transformation before the voxelized features are fed into the model. After the forward pass, we operate a corresponding inverse geometric transformation to the output features to ensure this representation shares the same properties with the non-transformed \textit{pathway-2}. Subsequently, we cluster features from the different point cloud blocks and produce two groups of clustering centroids and labels, which can be used for further training and loss assembled from different pathways.

\subsection{Superpoint} \label{sec:3.1}
\noindent
For all point clouds ${P_1, P_2, P_3, \ldots}$ within a point cloud set $\boldsymbol{P}$, we adhere to the VCCS\cite{VCCS} method to obtain initial superpoints for each point cloud. These can be denoted as $\left\{ \left\{SP_1^{1}, SP_1^{2}, SP_1^{3}, \ldots\right\}, \left\{SP_2^{1}, SP_2^{2}, SP_2^{3}, \ldots  \right\}, \ldots\right\}$, where $SP_j^{i}$ represents the $i$-th superpoint in the $j$-th point cloud. The initial superpoints may vary across different point clouds. Subsequently, we employ a straightforward strategy to merge the superpoints within each scene: 1) Identify the smallest superpoint $SP^{i}$ along with its two closest neighboring superpoints $SP^{k1}, SP^{k2}$; 2) Compute the vector addition of points within each superpoint and calculate the cosine similarity, here simply noted as $cos[SP^{i},SP^{k1}]$; 3) Merge the smallest superpoint with the one that exhibits higher cosine similarity; 4) Repeatedly execute the above three steps until the superpoints reach a predetermined number. This simplistic approach is based on the principle that similar semantic objects possess comparable normals. Ultimately, the updated superpoints become  $\left\{ \left\{SP_1^{n1}, SP_1^{n2}, \ldots\right\}, \left\{SP_2^{n1}, SP_2^{n2}, \ldots \right\}, \ldots\right\}$, ensuring that the points within the same superpoint are assigned identical labels. We define the final superpoint count as a parameter, represented by $\gamma_{sp}$. For all datasets, the optimal value is empirically found as  $\gamma_{sp} = 40 $.

\subsection{Voxelization and Devoxelization} 
\noindent  
We produce different representations for the input point cloud via the geometric transformation on the volumetric domain, where a voxel-based architecture is naturally adopted for such representation. Here, using voxelization and devoxelization in the pipeline, we present a simple yet effective network which contains only 3D convolutional layers with batch normalization without any additional component (details in \Cref{sec:3.3}).

Given the input points coordinate $\boldsymbol{c}$ with corresponding features $\boldsymbol{f}$ in the input blocks, we normalize the coordinates $\boldsymbol{c}$ before voxelizing the original points to gain scale invariance. Specifically, we normalize the coordinate $\boldsymbol{c}$ into [0,1] and denoted by $\boldsymbol{c}^{*} = \left\{c^{*}_{i}\right\}$. In this process, the point features (including the coordinates) do not change, and the normalized coordinates are only used for converting the feature to the proper volumetric space.

When transferring the features $\boldsymbol{f}$ with normalized coordinates $\boldsymbol{c^{*}} = \left\{ \boldsymbol{x^{*}},\boldsymbol{y^{*}},\boldsymbol{z^{*}} \right\}\ $ into the voxel grids $\left\{\boldsymbol{V}_{m,p,q}\right\}$, the interpolated feature $f_{i}$ for the voxel grid is calculated as the mean value of the features of points located in the grid.
\begin{equation}
\begin{split}
    \boldsymbol{V}_{m,p,q} = \frac{1}{K_{m,p,q}} \sum_{i=1}^{n}\boldsymbol{I} [ floor({x^{*}_{i}}\times {r}) = m,\\ floor({y^{*}_{i}}\times {r}) = p,floor({z^{*}_{i}}\times {r}) = q ]\times {{f}_{i}}
\end{split}
\end{equation}
where $r$ denotes the voxel resolution and $\boldsymbol{I}$ is an indicator function that indicates whether coordinates $c_{i}$ belong to the voxel grid ${ \left\{ m,p,q \right\} }$. ${K_{m,p,q}}$ represents the count of points falling within the grid ${\left\{ m,p,q \right\}}$, and $floor(\cdot)$ is floor function that outputs the greatest integer less than or equal to the input. 

In terms of the per-point clustering, we need to devoxelize the voxel-based features output from the model ${g_{\theta}}$ to point-based features. We follow the trilinear interpolation of PVCNN \cite{pvcnn} instead of the traditional nearest neighbor interpolation to ensure that nearby points are not assigned identical features.


\subsection{Baseline: Clustering and Iteration} 
\label{sec:3.3}
\noindent  
\ours applies a clustering-based method iteratively to generate pseudo-labels and train our baseline method, as inspired by DeepCluster \cite{deepcluster}. Adapting \cite{deepcluster} to the 3D domain is non-trivial due to the irregular nature and varying sparsity of point clouds. We present a simple yet effective strategy: switching between generating pseudo-labels via clustering with the current feature representations, and training new feature representations with the generated pseudo-labels. Different from \cite{deepcluster, picie}, the segmentation function ${g}_{\theta}$ should be able to produce per-point features, and we replace the parametric classifier with a non-parametric distance metric. Specifically, we denote the voxelization and devoxelization operations as ${\boldsymbol{Z}}$ and ${\boldsymbol{Z}^{-1}}$. The voxelized feature is $\boldsymbol{v}={\left\{ {v}_{i} \right\} } = \left\{ \boldsymbol{Z}({f}_{i},{c}_{i}^{*}) \right\}$, and the output voxelized feature of the 3D convolutional function is $\boldsymbol{v}^{out} = {g}_{\theta}(\boldsymbol{v})$. Finally, the features for clustering can be denoted as $\boldsymbol{f}^{'} = \left\{\ {f}_{i}^{'} \right\} = \left\{ \boldsymbol{Z}^{-1}({v}_{i}^{out},{c}_{i}^{*}) \right\}$. The main procedure can be separated as two parts:

(1) Using the current embeddings and \textit{k}-means to cluster the points with superpoints guidance in the point cloud:
\begin{equation}
\begin{split}
\min_{\boldsymbol{l,\mu }}\sum_{i} {\left \|{f}_{i}^{'}- \mu_{l_{i}^{sp}} \right \|}^{2}
\end{split}
\end{equation}
where $l_{i}^{sp}$ denotes the cluster label of point $c_{i}$ with the constraint of superpoint, and  $ \mu_{k}$ denotes the k-th cluster centroid. Note the features ${f}_{i}^{'}$ and the centroids $ \mu_{k}$ have the same dimension.

(2) Using the class labels as pseudo-labels, we train a classifier via cross-entropy loss, which is shown in the point cloud setting as:
\begin{equation}
\begin{split}
\min_{\theta,{W}}\sum_{i}\textit{L}_{CE}\left ( g_{W}({f}_{i}^{'}),l_{i}^{sp},\boldsymbol{\mu}  \right ) 
\end{split}
\end{equation}
where $g_{W}$ denotes the parametric classifier. Under the unsupervised setting, it will be very challenging to train a classifier jointly with constantly changing pseudo-labels. We therefore choose to label points only based on their cosine distance from to the clustering centroids in feature space. Specifically, the loss function shows the following format:
\begin{equation}
\begin{split}
\min_{\theta}\sum_{i}\textit{L}_{cluster}\left ( {f}_{i}^{'},l_{i}^{sp},\boldsymbol{\mu}  \right ) 
\end{split}
\end{equation}
\begin{equation}
\begin{split}
\textit{L}_{cluster}\left ( {f}_{i}^{'},l_{i}^{sp},\boldsymbol{\mu}  \right )=-log\left(  \frac{e^{-d({f}_{i}^{'},\mu_{l_{i}^{sp}})}}{\sum_{t} e^{-d({f}_{i}^{'},\mu_{t})}} \right)
\end{split}
\end{equation}
where $d(\cdot,\cdot)$ denotes the cosine distance.

\begin{figure*}
\begin{center}
\includegraphics[width=0.95\linewidth]{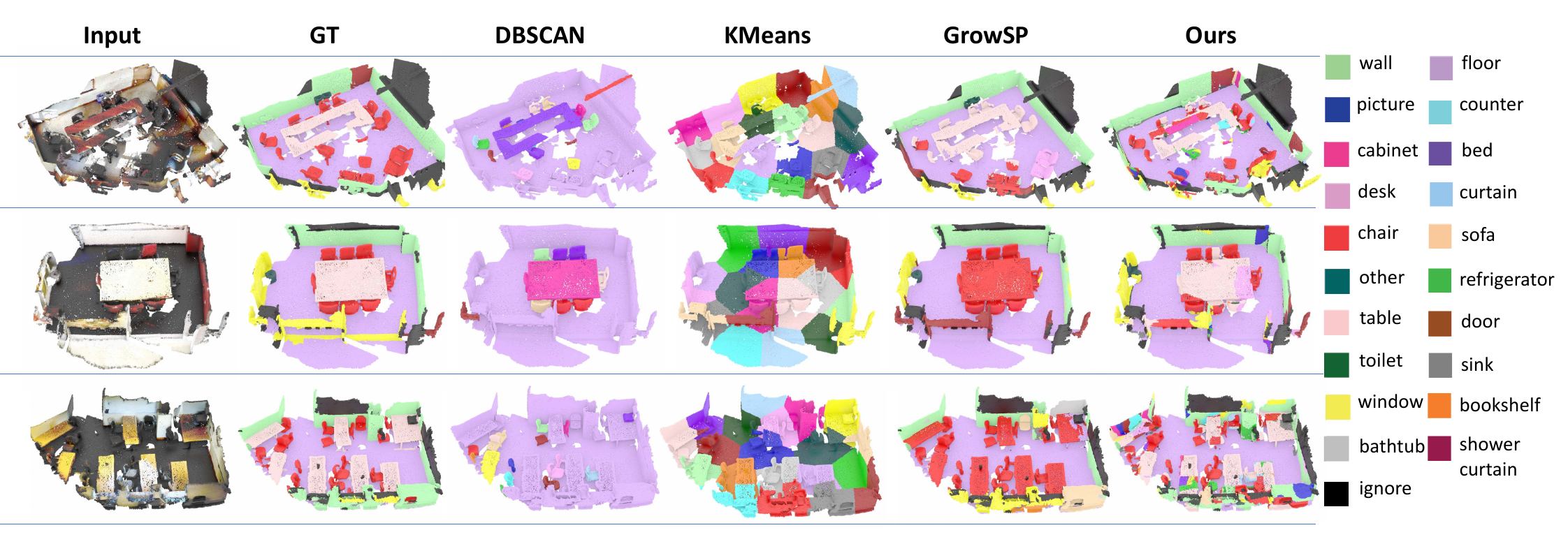} %
\end{center} 
   \vspace{-6mm}
   \caption{Qualitative results on ScanNet \cite{scannet}. Each class label is assigned a colour (as per legend, right). This illustration shows superior segmentation performance compared to the baselines.}
   \vspace{-2mm}
\label{fig:s3dis}
\end{figure*}


\subsection{Volumetric Transformations} \label{sec:3.4}
\noindent
To improve robustness in the unsupervised setting under different scenarios, we leverage the invariance and equivariance of volumetric representations of point clouds. Invariance means that the labelling should not change after applying different transformations such as colour jittering. Equivariance in the volumetric domain means when we apply a geometric transformation to the point cloud, the corresponding 3D convolutional feature should be similarly transformed, and the corresponding labels are also wrapped according to this transformation.

For simplicity, we name the two pipelines processing the two representations as \textit{pathway-1} and \textit{pathway-2}. To produce two different representations for an individual input block, we apply a geometric transformation before volumetric feature extraction and then perform a corresponding inverse transformation on the final voxelized features.

Specifically, let $\boldsymbol{G}$ and $\boldsymbol{G}^{-1}$ denote the voxelized feature geometric transformation and its reverse transformation respectively, and $\boldsymbol{O}$ is the colour transformation. For point $\boldsymbol{c}$ with its feature $\boldsymbol{f}$, we apply different colour transformations for original features $\boldsymbol{f}$:
 \begin{equation}
    \begin{split}
        \boldsymbol{f}_{1} = \boldsymbol{O}_{1}(\boldsymbol{f}) ,
        \boldsymbol{f}_{2} = \boldsymbol{O}_{2}(\boldsymbol{f})
    \end{split}
\end{equation}
Next, we transform these two features into the voxel grid, noting that $\boldsymbol{c}_{1}^{*}$ is actually equal to $\boldsymbol{c}_{2}^{*}$:
\begin{equation}
    \begin{split}
        \boldsymbol{v}_{1} = \boldsymbol{Z}(\boldsymbol{f}_{1},\boldsymbol{c}_{1}^{*}) ,
        \boldsymbol{v}_{2} = \boldsymbol{Z}(\boldsymbol{f}_{2},\boldsymbol{c}_{2}^{*})
    \end{split}
\end{equation}
After that, the voxelized feature transformations are applied to the volumetric domain:
only the features of \textit{pathway-1} are transformed whilst the other remains unchanged. 
The geometric transformations operate on the voxelized feature $\boldsymbol{v}$ and the corresponding reverse geometric transformations operate on the output voxel feature $\boldsymbol{v}^{out}$:
\begin{equation}
    \begin{split}
        \boldsymbol{v}_{1}^{out} = \boldsymbol{G}^{-1} \left\{  {g}_{\theta} [\boldsymbol{G} (\boldsymbol{v}_{1}) ] \right\} ,
        \boldsymbol{v}_{2}^{out} = {g}_{\theta}(\boldsymbol{v}_{2}) 
    \end{split}
\end{equation}
Subsequently, we perform de-voxelization to get the features for clustering:
\begin{equation}
    \begin{split}
        \boldsymbol{f}_{1}^{'} = \boldsymbol{Z}^{-1}(\boldsymbol{v}_{1}^{out},\boldsymbol{c}_{1}^{*}) ,
        \boldsymbol{f}_{2}^{'} = \boldsymbol{Z}^{-1} (\boldsymbol{v}_{2}^{out},\boldsymbol{c}_{2}^{*})
    \end{split}
\end{equation}


\subsection{Losses and Labelling Scheme}
\noindent  
Given input clouds $\boldsymbol{c}$ with features $\boldsymbol{f}$, according to the colour and geometric transformations introduced in \Cref{sec:3.3}, two different feature representations, $\boldsymbol{f}_{1}^{'},\boldsymbol{f}_{2}^{'}$, can be produced. By leveraging these two features, we cluster the two representations separately to get two groups of centroids and pseudo-labels:
\begin{equation}
\begin{split}
\boldsymbol{l^{(1)},\mu^{(1)}} = arg\min_{ \boldsymbol{l},\boldsymbol{\mu}}\sum_{i}\left \| {f}_{1i}^{'} -\mu_{l_{i}^{sp}} \right\|^{2}
\end{split}
\end{equation}
\begin{equation}
\begin{split}
\boldsymbol{l^{(2)},\mu^{(2)}} = arg\min_{ \boldsymbol{l},\boldsymbol{\mu}}\sum_{i}\left \| {f}_{2i}^{'} -\mu_{l_{i}^{sp}} \right\|^{2}
\end{split}
\end{equation}

We then set two loss functions. Firstly, the feature representation should match the pseudo-labels produced by the same pathway:
\begin{equation}
    \begin{split}
     \textit{L}_{1} =  \sum _{i}\textit{L}_{cluster}\left ( {f}_{1i}^{'},l_{i}^{^{sp}(1)},\boldsymbol{\mu^{(1)} }  \right ) +\\  
     \sum _{i}\textit{L}_{cluster}\left ( {f}_{2i}^{'},l_{i}^{^{sp}(2)},\boldsymbol{\mu^{(2)} }  \right )
    \end{split}
\end{equation}
Similarly, the feature representation should whilst match the pseudo-labels produced by the different pathway:
\begin{equation}
    \begin{split}
     \textit{L}_{2} =  \sum _{i}\textit{L}_{cluster}\left ( {f}_{1i}^{'},l_{i}^{^{sp}(2)},\boldsymbol{\mu^{(2)} }  \right ) +\\  
     \sum _{i}\textit{L}_{cluster}\left ( {f}_{2i}^{'},l_{i}^{^{sp}(1)},\boldsymbol{\mu^{(1)} }  \right )
    \end{split}
\end{equation}
The final training objective is their summation:
\begin{equation}
    \begin{split}
        \textit{L}_{final}=\textit{L}_{1} +\textit{L}_{2}
    \end{split}
\end{equation}

The loss encourages the feature from one pathway to adhere to labels generated by another pathway, which encourages the network to label similarly to feature representations from different pathways.

\textbf{Hungarian Algorithm}: To match the clustering labels with the real labels, we utilize the Hungarian algorithm \cite{HA} accross the whole dataset every epoch. Specifically, where $C$ is categories, $P$ is the predicted set and $G$ is the ground truth (GT) set. $S^{C\times C}$ is the matching matrix, where $S_{ij}$ denotes the matching degree between $i^{th}$ predicted category and $j^{th}$ GT category. Criterion: finding bijection ${\bf{f:}}i\rightarrow j$ to maximize $\sum_{i=1}^{C}S_{i,f(i)}$.


\section{Experiments}
\noindent  
\textbf{Implementation Details:} We implement a simple yet effective framework with 8 layers 3D convolution, where each layer employs a 3D batch normalization and leaky rectified linear activation function (ReLU). The input point cloud contains 12D features, i.e., the point coordinates $(x,y,z)$ in the normalized block coordinate system, colour information $(R,G,B)$, per-point normals and normalized raw coordinates in the original scene coordinate system. Note that no colour information is provided in SemanticKITTI\cite{behley2019semantickitti}. 

\begin{table}
\footnotesize
\begin{center}
\begin{tabular}{l|c|c|c|c}
\hline
Method & Level of Supervision  & mIoU & mAcc & oAcc \\
\hline
KMeans \cite{KMeans}  & unsupervised  & 3.4  & 10.4 & 10.2 \\
DBSCAN  \cite{DBSCAN}&  unsupervised & 6.1 & 10.1  & 15.3\\
GrowSP \cite{Unsup} &  unsupervised & 25.4 & 44.2  & 57.3 \\
\textbf{\ours (ours)}  &  unsupervised  & $\textbf{27.3}$ & $\textbf{46.8}$ & $\textbf{60.1}$ \\
\hline
\end{tabular}
\end{center}
   \vspace{-6mm}
\caption{Semantic segmentation results on ScanNet dataset. We evaluate 20 categories on validation set}\label{tab:scan}
   \vspace{-2mm}
\end{table}
\begin{figure*}
\begin{center}
\includegraphics[width=0.95\linewidth]{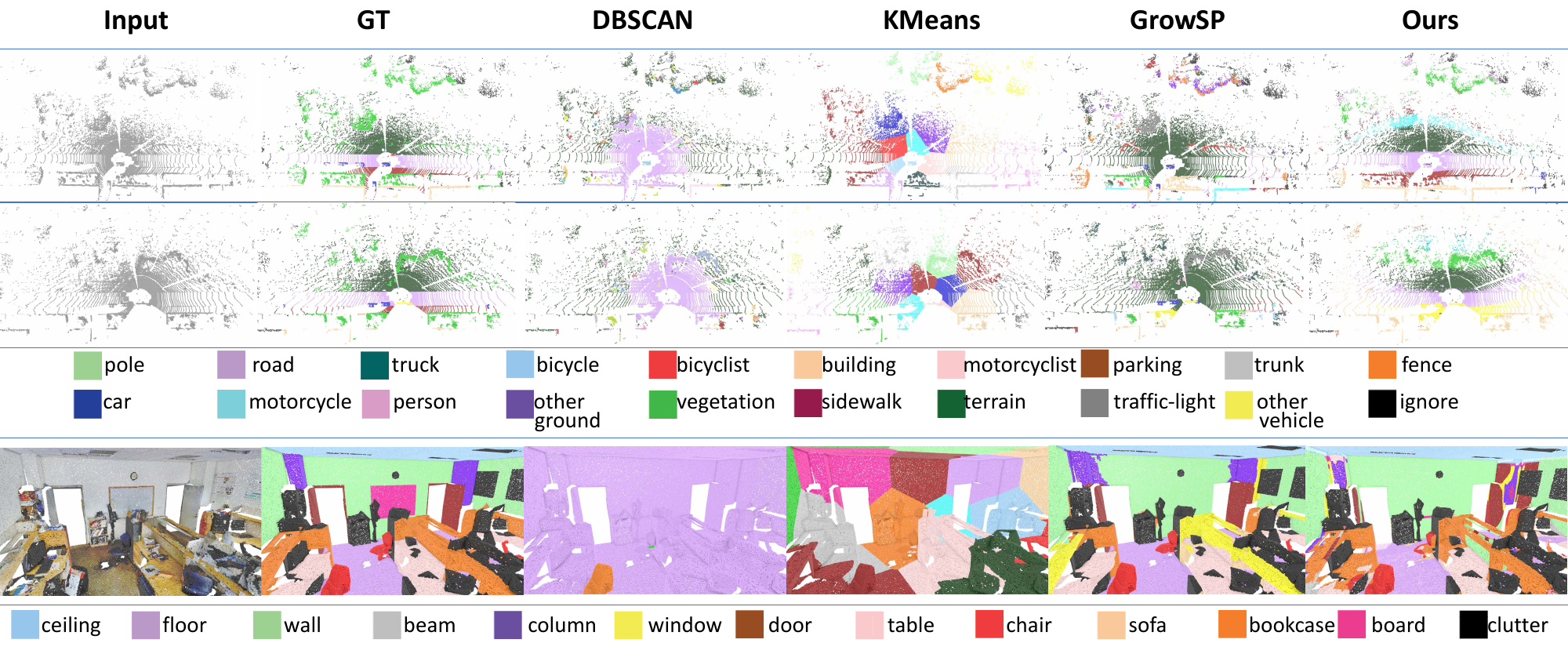}
\end{center} 
   \vspace{-7mm}
   \caption{Qualitative results on SemanticKITTI \cite{behley2019semantickitti} (top 2 rows) and S3DIS\cite{s3} (bottom row). 
   Our method draws more versatile results compared with DBSCAN \cite{DBSCAN} and is more stable than \textit{k}-means \cite{KMeans}, which shows promising segmentation results.}
   \vspace{-2mm}
\label{fig:semantic3d}
\end{figure*}
  
\textbf{Training:} We use a batch size of 4 with 4096 points per batch for all datasets. The chosen optimizer is stochastic gradient descent (SGD) with a learning rate of $1e-4$ and a  weight decay of $1e-5$. We train our network for 10 epochs. For the geometric transformation in the volumetric domain, we reverse the order of tensors along the given $x,y$ and $z$-axis respectively. The colour transformation comprises random contrast and random brightness adjustment. The output feature dimension from the model and the clustering feature dimension is set to 128. The resolution of the voxel grid is set to 32. Besides, we use the FAISS library \cite{faiss} on GPU to compute the cluster centroids via employing a mini-batch \textit{k}-means approach \cite{mini-batch}.

\textbf{Evaluation:} For evaluation and comparison with other methods, we choose two classical unsupervised clustering methods, \textit{k}-means \cite{KMeans}, DBSCAN \cite{DBSCAN}, and the only unsupervised semantic segmentation method GrowSP \cite{Unsup} as baselines. Our method is evaluated with three metrics: overall accuracy (oAcc), mean accuracy (mAcc) and the mean intersection of union (mIoU) on all datasets. All experiments are performed on a single NVIDIA RTX 2080Ti GPU.

\begin{table}
\footnotesize
\begin{center}
\begin{tabular}{l|c|c|c|c}
\hline
Method & Level of Supervision  & mIoU & mAcc & oAcc \\
\hline
KMeans \cite{KMeans}  & unsupervised  & 2.5 & 8.1 & 8.2  \\
DBSCAN  \cite{DBSCAN}&  unsupervised & 6.8 & 7.5& 17.8  \\
GrowSP \cite{Unsup} &  unsupervised & 13.2  & 19.7& $\textbf{38.3}$  \\
\textbf{\ours (ours)}  &  unsupervised  & $\textbf{14.2}$ & $\textbf{23.1}$& 34.8  \\
\hline
\end{tabular}
\end{center}
   \vspace{-6mm}
\caption{Semantic segmentation results on SemanticKITTI dataset. We evaluate 19 categories on validation set}\label{tab:kitti}
   \vspace{-2mm}
\end{table}

\subsection{Datasets}
\noindent  
We evaluate \ours on two indoor and one outdoor benchmark: S3DIS \cite{s3}, ScanNet \cite{scannet} and SemanticKITTI \cite{behley2019semantickitti}.

\textbf{S3DIS} \cite{s3} is a large-scale indoor scenes dataset which consists of 271 point cloud rooms in six areas. The annotations of each point in the point cloud scene belong to 13 semantic categories. We train the model in areas 1, 2, 3, 4, 6 and test it in area 5 following \cite{qi2017pointnet, pointcnn, kpconv}. We exclude clutter and test with 12 classes for a fair comparison with GrowSP\cite{Unsup}, nevertheless, we also test with 13 categories to compare with the existing supervised, weakly, and semi-supervised methods.

\textbf{ScanNet-v2 \cite{scannet}} is an RGB-D real-world indoor dataset. It contains 1201 scenes for training, 312 for validation, and 100 for online testing. For scene semantic segmentation, it has 40 classes and one unlabelled class for training and 20 classes and for testing. We compare with existing clustering and unsupervised methods on the validation set.

\textbf{SemanticKITTI \cite{behley2019semantickitti}:} is a large-scale outdoor dataset that is based on the KITTI Vision Odometry Benchmark. For the semantic segmentation task, it provides 22 sequences with point-wise annotation of 19 classes. Each sequence contains a number of scene scans collected by the complete 360 field-of-view of the employed automotive LIDAR, where sequences 11-21 are used for online testing, 08 is the validation set and the others are training sets. 

\textbf{Data Preparation:} For all datasets, we choose $\gamma_{sp}=40$ as the superpoint number for each scene. We first apply uniform downsampling to S3DIS\cite{s3} and ScanNet\cite{scannet} with the sub-grid size 0.03 and subsequently follow PointCNN \cite{pointcnn} to sample point clouds into blocks to ensure that each data sample in the batch has the same number of points. For S3DIS \cite{s3} and ScanNet\cite{scannet}, the block size is $1.5 \times 1.5 $ on $xy$ plane, and each block contains 4096 points. For SemanticKITTI \cite{behley2019semantickitti}, we set each block size as $5 \times 5 $ on $xy$ plane with 4096 points. For each point cloud, we utilize VCCS\cite{VCCS} to derive the initial superpoint. This is then merged for enhanced segmentation, as detailed in \Cref{sec:3.1}. Furthermore, due to the characteristics and predominance of roads in outdoor SemanticKITTI\cite{behley2019semantickitti} datasets, we apply RANSAC\cite{RANSAC} to fit a plane as the road for improved generation of superpoints. Note this process will not be utilized elsewhere. 

\begin{table}
\footnotesize
\begin{center}
\resizebox{0.47\textwidth}{!}
{\begin{tabular}{l|c|c|c|c}
\hline
Method & Level of Supervision  & mIoU & mAcc & oAcc \\
\hline
PTv2 \cite{PTV2} & fully supervised   & 72.6  & 78.0&91.6 \\
KPConv \cite{kpconv} &fully supervised  & 67.1 & 72.8& - \\
SSP+SPG \cite{SSP} & fully supervised& 61.7  &68.2& 87.9  \\
PointNet \cite{qi2017pointnet} & fully supervised& 41.4 & - & -\\
\hline
\hline

Jiang et al. \cite{semi}  & semi-supervised ($10\%$) & 57.7  & - &69.1\\
\hline
MT \cite{MT} & weakly supervised (1pt) & 44.4  & - &- \\
Zhang et al. \cite{weakly} &weakly supervised (1pt) & 48.2 & - &- \\
\hline
\hline
KMeans \cite{KMeans}& unsupervised & 9.4 & 21.2 & 22.1 \\

DBSCAN \cite{DBSCAN} & unsupervised  & 9.2 & 19.8 & 17.5 \\

GrowSP(12) \cite{Unsup}  & unsupervised  & \textbf{44.6} & \textbf{57.2} & \textbf{78.4} \\

\ours (ours)(12) & unsupervised  & 42.8& 55.8 &75.5
\\
\ours (ours) & unsupervised  & 40.1& 52.9 & 72.3

\\
\hline
\end{tabular}}
\end{center}
   \vspace{-6mm}\
\caption{
Semantic segmentation results on S3DIS Area-5 are compared using mIoU, mAcc and oAcc across various methods. Where (12) indicates the exclusion of clutter, while the results without (12) are tested with 13 classes.
}\label{tab:s3dis}
   \vspace{-2mm}
\end{table}

\subsection{Results and Comparison on Benchmarks}
\noindent  
To thoroughly evaluate our \oursnospace, we test our methods on the indoor S3DIS \cite{s3}, ScanNet\cite{scannet} and outdoor SemanticKITTI\cite{behley2019semantickitti} benchmarks. \cref{tab:scan,tab:kitti,tab:s3dis} respectively shows the semantic segmentation results on the ScanNet, SemanticKITTI and S3DIS dataset. Not surprisingly, fully supervised methods provide the best performance. From \cref{tab:s3dis}, our method significantly outperforms the existing clustering methods, where it achieves 75.5\% overall accuracy and 42.8 mIoU on the S3DIS dataset. Moreover, our method is even close to the performance reported by the fully supervised method \cite{qi2017pointnet} and some up-to-date weakly supervised methods \cite{weakly2, MT}, which is a big step forward for unsupervised semantic 3D scene segmentation.

Moreover, we outperform GrowSP \cite{Unsup} on both the ScanNet and SemanticKITTI datasets. Specifically, as displayed in \cref{tab:scan}, our method achieves a superiority of +1.9 mIoU and +2.6 mAcc over their results. Additionally, \cref{tab:kitti} demonstrates that our method achieves 1 mIoU and 3.4 mAcc higher than GrowSP \cite{Unsup}, despite having a slightly lower oAcc. \cref{fig:s3dis} shows the qualitative comparison on S3DIS, which further demonstrates the superiority of our method. 

\subsection{Ablation Study} 
\noindent  
To showcase the effectiveness of each module and the different volumetric transformations. We conduct eight groups of experiments on the S3DIS \cite{s3} dataset: (1) the baseline approach proposed in \Cref{sec:3.3}, (2) adding colour transformation on the basis of the control group (1), (3) adding voxelized feature transformation on the basis of the control group (1), and (4) full model without prior superpoint, (5)-(8) different final prior superpoints as guidance. As shown in \cref{tab:ablation}, our full model clearly outperforms the baseline on all of the evaluation metrics, benefiting from the delicate volumetric transformation design and superpoint prior. Groups (3) and (4) outperform by \textbf{+5} mIoU and \textbf{8} OA compared to the baseline. More interestingly, the improvement of adding the geometric transformation for equivariance is more significant than that of the invariance transformations, which is different from prior unsupervised learning work in the 2D domain \cite{2dULGans,2dULpartseg,2DULcontrasmask}. It is known that point clouds essentially present much stronger geometric priors than 2D images with explicit 3D structures, which we believe can significantly help the 3D representations to be more robust and consistent cross-view and less sensitive to light changes and jittering. Moreover, the employment of superpoints can significantly enhance the overall performance. This enhancement is a result of the more abundant information of superpoints, which facilitates the pre-segmentation of the scene into higher-level semantic classes. Additional results are available in the supplementary material. 

\begin{table}
\footnotesize

 \begin{center}
    \begin{tabular}{ccccccc}
    \hline 
       Baseline   & Eqv       & Inv       & $\gamma_{sp}$ & mIoU       & mAcc     & oAcc \\
    \hline
      \checkmark  &           &           &    & 29.8       &   42.5      & 55.3  \\
      \checkmark  &           &\checkmark &    & 30.7       &   43.5         & 57.2 \\
      \checkmark  &\checkmark &           &    & 33.9       &     45.9      & 61.4 \\ 
      \checkmark  &\checkmark &\checkmark &    &    34.8        &  46.3     &   63.2        \\
      \checkmark  &\checkmark &\checkmark & 80   &   38.8       &  49.7    &    68.7       \\
      \checkmark  &\checkmark &\checkmark & 60   &   41.0       &   52.6    &   72.4        \\
      \checkmark  &\checkmark &\checkmark & 40   & \textbf{42.8}& \textbf{55.8}      & \textbf{75.5} \\
      \checkmark  &\checkmark &\checkmark & 20   &    41.9     &   53.9    &   74.3        \\
      
    \hline
    \end{tabular}
    \end{center}
   \vspace{-6mm}
    \caption{Ablation study on S3DIS Area-5: Eqv denotes equivariant voxelized feature transformation; Inv denotes invariant colour transformation. $\gamma_{sp}$ denotes the final 
    superpoint number.}\label{tab:ablation}
   \vspace{-2mm}

\end{table}

\subsection{Analysis}\label{sec:analysis}
\noindent 
Our \ours approach demonstrates a promising level of performance on both indoor and outdoor datasets when compared to existing baselines. In contrast to GrowSP\cite{Unsup}, our method achieves superior results on ScanNet\cite{Unsup} and SemanticKITTI\cite{behley2019semantickitti}. As the scene complexity increases, the quality of GrowSP\cite{Unsup} superpoints tends to degrade. In contrast, our approach not only incorporates pre-segmentation but also employs a two-pathways training algorithm, leveraging the concepts of invariance and equivariance. 

Nonetheless, slight performance degradation can occur in practical scenarios. To address this, we have implemented three strategies: (i) splitting the largest cluster when another cluster in the set reaches zero entities; (ii) applying mild centroid perturbation during updates; and (iii) re-weighting for loss balancing using per-class pseudo-label ratios at each epoch. Additionally, our two-pathways approach expedites the convergence time during training. For instance, while training with only one pathway necessitates around 8 epochs to achieve convergence, the two-pathways approach accomplishes convergence in just 2-3 epochs.



\section{Conclusion and Discussion}
\noindent  
We propose a novel generalized unsupervised semantic segmentation method for both indoor and outdoor 3D scenes with objects and the background. Our method leverages a simple yet effective framework via clustering and iterative generation leveraging the invariance and equivariance of the volumetric representations with the assistance of superpoint. Experiments show promising performance on S3DIS, ScanNet and SemanticKITTI datasets which proves the superiority of our approach beyond all the existing baselines. This work aims to provide more insight for 3D unsupervised learning. Future work will explore improved point sampling strategies and an extension to point- or graph-based representations, benefiting other areas related to unsupervised learning, metric learning and 3D representation learning.
  

\noindent
\textbf{Acknowledgement:} EPSRC NortHFutures (ref: EP/X031012/1).

{\small
\bibliographystyle{IEEEbib}
\bibliography{egbib}
}

\end{document}